\documentclass[journal]{journal}
\usepackage[colorinlistoftodos,prependcaption,textsize=tiny]{todonotes}
\usepackage{amsmath}

\ifCLASSINFOpdf
\else
\fi

\begin{document}
\title{ViraPart: A Text Refinement Framework for Automatic Speech Recognition and Natural Language Processing Tasks in Persian}

\author{Narges Farokhshad, Milad Molazadeh, Saman Jamalabbasi, Hamed Babaei Giglou, Saeed Bibak
\thanks{Authors are with Part AI Research Center, NLP Department, Tehran, Iran (e-mails:\{narges.farrokhshad, milad.molazadeh, saman.abbasi, hamed.babaei, saeed.bibak\}@partdp.ai)}}

\markboth{Journal of \LaTeX\ Class Files,~Vol.~6, No.~1, January~2007}%
{Shell \MakeLowercase{\textit{et al.}}: Bare Demo of IEEEtran.cls for Journals}

\maketitle
\thispagestyle{empty}

\begin{abstract}
The Persian language is an inflectional subject-object-verb language. This fact makes Persian a more uncertain language. However, using techniques such as Zero-Width Non-Joiner (ZWNJ) recognition, punctuation restoration, and Persian Ezafe construction will lead us to a more understandable and precise language. In most of the works in Persian, these techniques are addressed individually. Despite that, we believe that for text refinement in Persian, all of these tasks are necessary. In this work, we proposed a ViraPart framework that uses embedded ParsBERT in its core for text clarifications. First, used the BERT variant for Persian following by a classifier layer for classification procedures. Next, we combined models outputs to output cleartext. In the end, the proposed model for ZWNJ recognition, punctuation restoration, and Persian Ezafe construction performs the averaged F1 macro scores of 96.90\%, 92.13\%, and 98.50\%, respectively. Experimental results show that our proposed approach is very effective in text refinement for the Persian language.
\end{abstract}

\begin{IEEEkeywords}
Persian Ezafe, Punctuation, ZWNJ, NLP, ParsBERT, Transformers

\end{IEEEkeywords}
\IEEEpeerreviewmaketitle

\section{Introduction}

\IEEEPARstart{M}{ost} of the spoken languages around the world such as Persian are rhythmic and this fact will make the words of sentences more meaningful while speaking and one can change the meaning of a sentence by changing the purpose and significance of a particular word.  By making sure that marks are used accurately, we can avoid probable uncertainty. Otherwise, the sentence will be understood differently; hence, not paying attention to the accurate use of punctuation leads to confusion and misunderstandings \cite{persian:rythm}.

Automated Speech Recognition (ASR) is a technology that allows users to speak entries rather than punching numbers or letters on a keypad. However, in a typical ASR system, punctuation and capitalization of words are removed because they do not affect the pronunciation of words. The output contains purely a sequence of words or alphabets characters. This output is sufficient for many applications that usually use a short and independent segment of speeches, but it is difficult to be used in applications that decipher long speech segments. Furthermore, ASR results are fed into natural language processing (NLP) models that punctuations and word capitalizations are important pieces of information that can help to boost the NLP model's performances \cite{cho:segment2020}.

To achieve a readable system in Persian there are different tasks. We considered punctuation restoration, zero-width non-joiner (ZWNJ) recognition, and Ezafe construction tasks to process texts for NLP tasks. \textit{Punctuation restoration} is the task of identifying punctuation marks that increases the readability of texts. Punctuation restoration and correcting word casing is one requirement to offer a reliable post-processing system for texts before applying complex NLP algorithms. For English, we obtained comparable state-of-the-art results, while for Persian, there are a limited number of works in the field. In the English language, hyphens are used to join words and to separate syllables of a single word. This process is called hyphenation and this type of word is called multi-part word. However, the Persian language consists of multi-part words as well. In Persian morphology, the \textit{ZWNJ} character is used to separate parts of multi-part words \cite{zahedi2016}. 
Persian ezafe is an unstressed morpheme syllable that appears at the end of the words. It is pronounced as -e after consonants and as -ye after vowels. This syntactic phenomenon links a head noun, head pronoun, head adjective, head preposition, or head adverb to their modifiers in a constituent called \textit{ezafe construction} \cite{Nassajian2020}.

In most studies, all of these three concepts are investigated separately (Punctuation restoration, ZWNJ recognition, and ezafe construction) one reason could be the avaliable dataset that are for different purposes. Another reason is the complexity of the tasks is varied from one task to another. However, in this work, we analogy that for an enriched text we must pay attention to all three aspects, so in this study, we proposed a framework that is a combination of three models for text refinement. In practice, we used publically avaliable datasets to create models that make the readable text for numerous NLP tasks. In general our research contributions are two folds:
\begin{itemize}
    \item We proposed a framework for text refinements in Persian for various Nlp tasks. We used three major text clarifications models in one framework.
    \item We used ParsBERT (a BERT variant for the Persian language) for sequence labeling. 
\end{itemize}
After reviewing the related works of three tasks in Section II, we introduce our proposed method in Section III. We then discuss punctuation restorations, ZWNJ recognition, and ezafe recognition tasks and their results in Sections IV and at the end, we made a conclusion of this paper in Section V.

 




\section{Related Works}
We first review methods for English, and then methods for the Persian language. Since our work relies on multiple tasks, we summarize related literature on these topics as well.

\subsection{English}
There are lots of works in the English language, we divided the related work in the field into 4 categories namely prosodic features, n-gram language model, RNN based approaches, and transformers. Prosodic features are features that appear when we put sounds together in connected speech. The \cite{Christensen2001} used a statistical finite-state model that combines prosodic, linguistic, and punctuation class features for the punctuation restoration task. In a similar work, the \cite{Kim2003} proposed a combined punctuation generation and speech recognition system that uses prosody features to boost their performance. The n-gram language models (LM) \cite{Beeferman1998} have a long history in NLP tasks that are useful for punctuation annotation systems. The \cite{Liu2006} proposed a metadata detection system based on n-gram LM that combines information from different types of textual knowledge sources with information from a prosodic classifier to automatic detection of sentence boundaries and disfluencies. In \cite{Gravano2009} authors used n-gram LM and larger training datasets consistently, and they conclude that using large data will improve performance while increasing n-gram order does not help. With progress in deep learning models, RNN based models achieved higher attention in the text-domain. Subsequent research revealed different approaches to this task. For example, authors of the \cite{Zelasko2018} proposed Deep Neural Network (DNN) sequence labeling model that uses Bidirectional Long Short-Term Memory (BiLSTM) and a Convolutional Neural Network (CNN) to predict the punctuation. The word and character-level embedding with CNN-RNN model for punctuation restoration excavated in \cite{Tundik2018}. LSTM approach has been proposed by \cite{Xu2016} authors for the punctuation prediction task. \\
The transformer in NLP is a novel architecture that aims to solve sequence-to-sequence tasks while handling long-range dependencies with ease \cite{attention}. Transformers are groundbreaking models in various NLP tasks such as sequence labeling. In works of \cite{Lin2020}, \cite{Yi2020}, \cite{Courtland2020}, \cite{Nagy2021}, and \cite{Makhija2019} authors used transformers models for punctuation restoration task and achieved a promising results.

\subsection{Persian}
In this section, we divided the related works into three sections, to explore similarly.

\subsubsection{ZWNJ Recognition} ZWNJ has been investigated in Persian and there are a few related works for this task. To the best of our knowledge, the oldest work on the shelf belongs to \cite{zwnj3}, in which they proposed a model based on a statistical machine translation paradigm for space corrections. In machine translation, the text in the source language is translated into a text in the destination language, in the proposed model the output is space corrected text. The BLEU score statistical machine-translation method reaches 0.91. Parsivar \cite{zwnj2} is a toolkit that performs different kinds of activities composed of normalization, space correction, tokenization, stemming, parts of speech tagging, and shallow parsing. They proposed a Naïve Bayes model for space correction with a BIO scheme to find words with multiple parts separated by spaces with an F1 score of 89.50\%. In a similar manner authors of \cite{zwnj5} investigated various nlp toolkits for the Persian language and they found out that except Parsivar none of them is capable of handling spaces desirably than Parsivar. They believe that Hazm\cite{hazm} is not properly handling the spaces in Persian, so they trained a model that uses Hazm tokenization as input for the space correction model. First, they combined various datasets and achieved 1,750,607 words. Next, they trained the n-gram language model using the KenLM \cite{kenlm} toolkit for space correction. Their method outperforms the Hazm model for space correction with an F1 score of 81.94\%. The BERT-based approach proposed by \cite{zwnj11} for word segmentation correction and ZWNJ recognition. They approach the problem jointly as a sequence labeling problem and proposed a BERT-CRF model which consists of a BERT model followed by a CRF layer for sequence labeling task for ZWNJ recognition. They specified three variant classes namely none, space, and ZWNJ classes. Next, They proposed an algorithm that extends the existing dataset by adding noses. This allowed boosting their model performance from an F1 score of 96.67\% to 98.14\%. However, they collected 500 difficult sentences for further evaluation and reported an F1 score of 92.40\%.
\subsubsection{Punctuation Restoration}
Punctuation prediction/restoration is a crucial task in ASR. The sequence of words with no punctuation needed to be processed to make sense for humans and even other areas of NLP. The \cite{Hosseini} introduced the first-ever corpus for automatic punctuation prediction in Persian texts. They have made many revisions to their corpus such as word replacements, normalizations, word type corrections, and numerous corrections to the punctuation marks. They trained the CRF model and achieved an F1 score of 69.00\% in their preliminary experiments.
\subsubsection{Ezafe Construction}
Ezafe construction is a peculiar aspect of the Persian language which is able to make a good indicator of important information in the text.  The \cite{ezafe4} mentioned that adding information about ezafe can boost performances of dependency parsing and shallow parsing by 4.6\% and 9\% respectively. Most of the work in the field used rule-based and hand-crafted features to tackle this challenge. The \cite{muller} proposed the head-driven phrase structure grammar to formalize Persian syntax and determine phrase boundaries and \cite{megerdoomian} used a rule-based method to create a Persian morphological analyzer. The \cite{koochari} employed the CART tree classification approach for ezafe constructions. The \cite{asghari} used a probabilistic approach namely CRF to create an ezafe recognition model in Persian and achieved accuracy 98.04\%. Another work is \cite{shamsfard} that uses both a rule-based method and a genetic algorithm for ezafe recognition and reported accuracy of 95.26\%.  The newest work in ezafe recognition is the work of \cite{ezafe1} that uses transformers and ezafe recognition role in part-of-speech tagging. They employed transformer-based methods, BERT and XLMRoBERT, and achieved an F1 score of 98.10\%.

\section{Proposed Method}
\begin{figure}[!t]
\centering
\includegraphics[width=3.5in, height=4in]{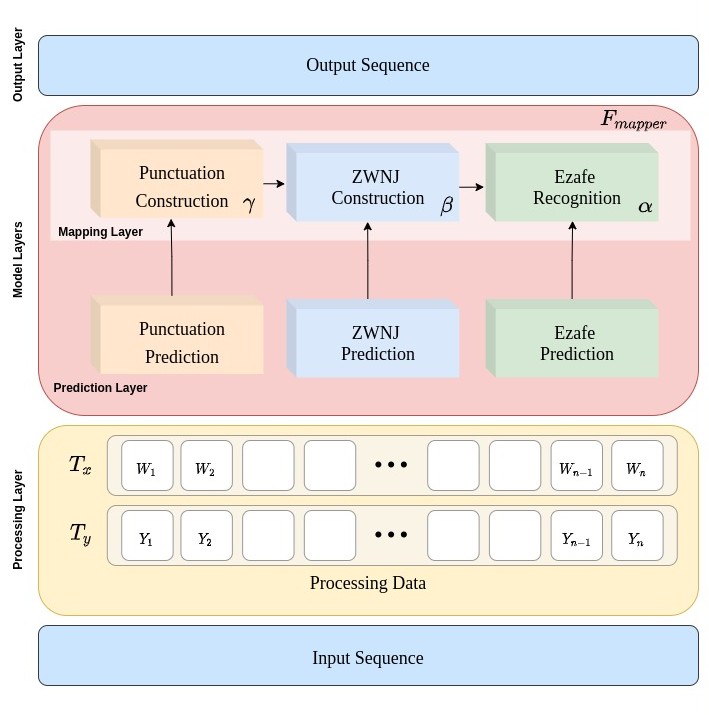}\\
\caption{Proposed System Architecture}
\label{fig_general_diag}
\end{figure}

In this section, the methodology of the proposed system is described elaborately. The proposed system consists of four layers including the input processing layer, model layers which are divided into two modules (prediction and mapping layers), and the output construction layer. The architecture of the proposed method is presented in figure \ref{fig_general_diag}. First, in the processing layer documents will be cleaned, normalized, and tokenized for the next layer. Next, the model layer that consists of prediction and mapping layers uses processed documents for punctuation construction, ZWNJ construction, and ezafe recognition. In the last layer, outputs will be combined for the final result of the system. The model layers consist of three sections, each for a different task that works in parallel form. The prediction layer in model layers tries to train models in different sections; in the punctuation section it tries to identify the possibility of specified punctuation marks. The ZWNJ section at the prediction layer predicts whatever ZWNJ should occur or not. Similarly, it works in the ezafe section in this layer as well. The next layer in model layers is the mapping layer which tries to map the prediction results on text. Mapping starts with the punctuation section by constructing punctuations. Then the produced output of the punctuation section feeds into the ZWNJ section for ZWNJ construction at the mapping layer. Likewise, the punctuation and ZWNJ constructed output from the previous section will feed into the ezafe section to determine which token needs ezafe construction. In the following sections, we will describe each layer more elaborately.

\subsection{Problem Formulations}
In this paper, our training data consists of $N$ sentences that each sentence consists of $n$ tokens. Regarding this; we can define $T_X$ and $T_Y$ as input and output representation of input sequences, respectively. The definitions are presented as follows:

$$ T_X = \sum_{i=1}^{n}W_i \;\;\&\;\; T_Y = \sum_{i=1}^{n}Y_i$$

Where $T_y$ changes based on models. In this paper, we defined the $T_Y$ for each model at the model layer as followings:

$$T_{Y_{punct}} = \sum_{i=1}^{n}Y_i \;\;\;  where \;\; Y_i \in C(".", ":", ",", "?", "unk" )$$
$$T_{Y_{zwnj}} = \sum_{i=1}^{n}Y_i \;\;\;  where \;\;\; Y_i \in C(1, 0)$$
$$T_{Y_{ezafe}} = \sum_{i=1}^{n}Y_i \;\;\; where \;\;\; Y_i \in C(1, 0)$$

Where $C$ is the label set for tokens in different tasks. For example, for punctuation tasks, we specified only four types of marks for restoration and the last \textit{unk} label is for a token that doesn't take any punctuation mark, in our model it is a five-class sequence labeling model. For other tasks, it is a binary class classification that 1 means changes required else changes are not required. Changes for binary classification are construction and specification (recognition) for ZWNJ and Ezafe tasks respectively. Overall, we are dealing with a classification task. Considering definitions, we can define the training, validation, and test sets as a following for each problem.

$$T_{train} = \{(T_{X_1}, T_{Y_1}), (T_{X_2}, T_{Y_2}), ...,  (T_{X_n}, T_{Y_n})\}$$
$$T_{val} = \{(T_{X_1}, T_{Y_1}), (T_{X_2}, T_{Y_2}), ...,  (T_{X_k}, T_{Y_k})\}$$
$$T_{test} = \{(T_{X_1}, T_{Y_1}), (T_{X_2}, T_{Y_2}), ...,  (T_{X_m}, T_{Y_m})\}$$

Where $n$, $k$, and $m$ are sizes of the train, validation, and test sets respectively. The goal is to create system $\gamma$, $\beta$, and $\alpha$ models that $F_{mapper}$ uses for punctuation, ZWNJ, and ezafe constructions.

$$F_{mapper}(T_X) = \alpha(\beta(\gamma(T_X)))$$

Where $\gamma$ is the punctuation construction model that predicts and constructs the specified punctuation marks in input $T_X$, and the resulting output $\hat{T_{Y}}$ will be used in the $\beta$ model for the joining of ZWNJ's. In the end, the output of the previous model will enter into $\alpha$ model for ezafe construction. 

\subsection{Processing Layer}
Raw documents first enter into the processing layer, the processing layer aims to create $(T_X, T_Y)$ pairs from input raw documents. Generally speaking, raw documents can be sentences or paragraphs as well and they need to be processed into the form of input pairs for models. Nevertheless, in this layer, we removed every punctuations mark from inputs. This technique will allow us to use the proposed framework for punctuation restoration tasks. It presents everything similar to the output of ASR systems. However, it is usable for punctuation compilation as well. 

First, a document is converted into $W_i$ tokens using a personalized tokenizer that considers spaces, half-spaces, and punctuations for splitting raw documents into texts. Next, the special character has been erased from $W_i$ tokens. Next, English numbers have been converted into Persian numerics. In the end, modifications on $Y_i$ applied to output required labels for $W_i$ sequences.
\subsection{Model Layers}
Model layers aim to make a prediction and mappings for refinements. In these layers, a model will be trained in the prediction layer to predict new unseen input sequences and the mapping layer will use the trained model for refinements. In the following, we will describe each layer carefully. 

\begin{figure}[!t]
\centering
\includegraphics[width=3.3in, height=3in]{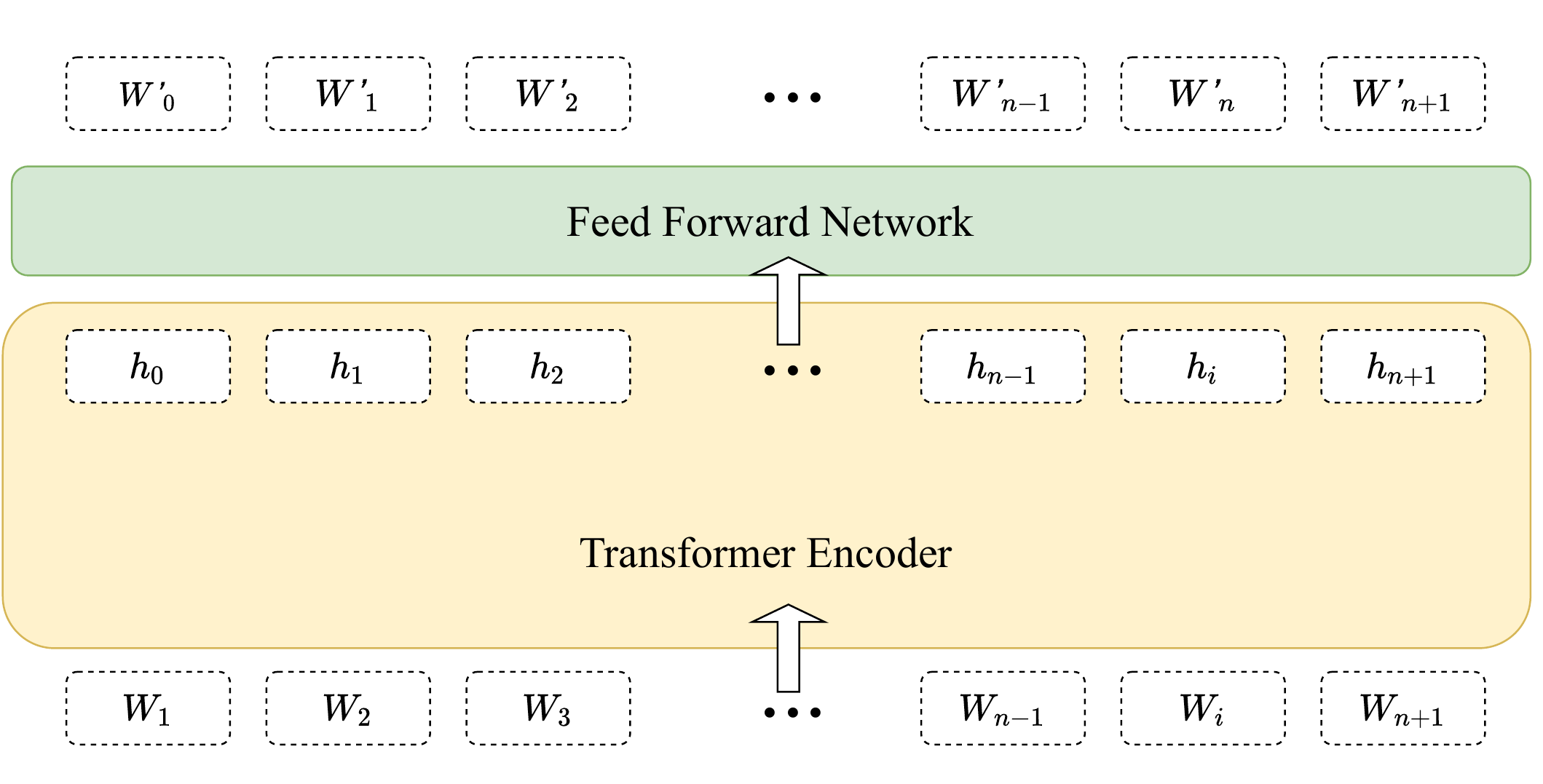}\\
\caption{Proposed Model Architecture}
\label{fig_model}
\end{figure}

\subsubsection{Prediction}
ParsBERT \cite{parsbert} is a monolingual version of the BERT language model \cite{bert} for the Persian language that uses the base configuration of the BERT model such as 12 hidden layers, hidden size of 768 with 12 attention heads. ParsBERT is a transformer-based language model with an encoder-only structure. The input sequence ${W_1, W_2, ..., W_n}$ is mapped to a contextualized encoded sequence ${W^{'}_1, W^{'}_2, ..., W^{'}_n}$ by going through a series of bi-directional self-attention blocks with two feed-forward layers in each block. Then we took the output sequences and mapped them to a binary and multi-classification with sigmoid and softmax activation functions for each task, respectively. First, we used ParsBERT tokenizer to create attention mask and word mappers, then we fine-tuned ParsBERT for tasks. For the training procedure, we used \textit{CrossEntropy} and \textit{CategoricalCrossEntropy} loss functions for binary and multi-class classification models, respectively. The rest of the parameters are the same for all of the models. We set, learning rate to \textit{2e5}, epoch to \textit{3}, and dropout to \textit{0.1} to train models.

\subsubsection{Mapping}
The mapping goal is to use the trained model in the prediction layer for making changes into tokens. The $F_{mapper}$ takes raw $T_X$ as input and then first uses the punctuation prediction model to identify which punctuation mark is needed to be applied for tokens, after punctuation construction ($\gamma$), the refined $T_X$ enters into the ZWNJ construction ($\beta$) for adding changes. Similarly, the output of $\beta$ will enter into ezafe recognition ($\alpha$) to sign ezafe in predicted tokens.

\subsection{Output Layer}
The output layer at top of the proposed framework combines the previous layer predictions from the mapper to construct the sentences from output sequences. 

\begin{figure}[!t]
\centering
\includegraphics[width=3in]{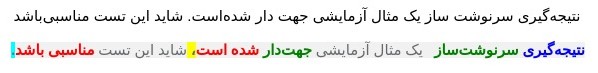}\\
\caption{Output layer result for raw input sentences.}
\label{output_layer}
\end{figure}

The figure \ref{output_layer} represents sample input and outputs from the proposed framework.  In this figure, the blue-colored text presents the correct ZWNJ constructed in input sentences and the model do not need to make any changes on them. However, the green-colored and red-colored texts are wrong ZWNJ construction in input sequences that our models recognized and corrected in the outputs. In general, there are two type of modifications for ZWNJ; the first type is occurring in green-colored texts, where ZWNJ is required in input sentences, the second type is occurring in the red-colored text, where ZWNJ is not required in input sentences and our model removes them in outputs. Punctuation mark restorations in texts happened with background-colored punctuations where yellow color means that it was incorrectly has been utilized in input sentences and blue color means that input sentences missing those punctuations. More experimental evaluations will be presented in the next sections.


\section{Results}
In this section, we will explore the dataset and results of the proposed framework.
\subsection{Dataset}
For training and evaluation of the proposed framework, we used the Bijankhan corpus \cite{Bijankhan}. The Bijankhan is a well-known dataset for Persian natural language processing tasks. The dataset consists of news articles and common texts consisting of in total 4300 different topics. This dataset consists of 2,597,939 words that are annotated with morpho-syntactic and partly semantic features. The corpus has been tagged by 550 tags in hierarchical order with more fine-grained POS tags like \textit{noun-plural-subj}. Nearly 23\% of words in this corpus are tagged with Ezafe. Punctuation marks are specified with the \textit{DELM} tag in this corpus with a total number of 246,023. Each word in this corpus is specified as a single row and sentences divided by \#. Multi-words are also specified in a single row, so in this way, we made modifications to the dataset using its structure to obtain a dataset for the ZWNJ model. We split multi-words into separate words and labeled them for ZWNJ recognition. We found almost 30\% of the corpus appropriate for the ZWNJ recognition task. We split the dataset for each task in 80\% 10\% 10\% for train, validation, and test sets, respectively. The table \ref{tab:dataset} presents the details about train, test, and validations sets based on tasks.

\begin{table}[]
    \centering
     \caption{The number of sentences and tokens in train, validation, and test sets.}
    \begin{tabular}{c|c|c}
         \hline
         Set & \# of Sentences & \# of Tokens \\
         \hline
         Punctuation\\
         \hline
         Train &  72780 &  2,040,213\\
         Validation & 9098 & 256,767\\
         Test  & 9097 & 255,197\\
         \hline
         Ezafe\\
         \hline
         Train & 78741 & 2,068,316\\
         Validation & 9843 & 256,282 \\
         Test  & 9843 & 261,362 \\
         \hline
         ZWNJ\\
         \hline
         Train & 75562 & 2,211,741 \\
         Validation & 9445 & 279,284 \\
         Test  & 9445 & 276,685\\
         \hline
    \end{tabular}
   
    \label{tab:dataset}
\end{table}

\subsection{Metrics}
TThe model's evaluation metrics are precision, recall, F1, and accuracy measures. Because the dataset is skewed for tasks, Macro $F1$ measurement provided a better view of the proposed method performance over all classes. For each class $C$, we calculate its precision ($p$), recall ($r$), and $F1$ score as follow:

$$p = \frac{\#\;of\;tokens\;predicted\;as\;C\;accurately}{\#\;of\;tokens\;predicted\;as\;C}$$
$$r = \frac{\#\;of\;tokens\;predicted\;as\;C\;accurately}{\#\;of\;tokens\;with\;grand\;truth\;label\;of\;C}$$

$$F1 = 2*\frac{p * r}{p + r}$$

Considering all the classes together, the averaged F1 macro score is:

$$Macro\;F1 = \frac{1}{n_c}\sum^{i=1}_{n_c}F1_i$$

Where $n_c$ is the number of classes, and $F1_i$ is the averaged F1 macro score for class $i$. The overall accuracy for all tokens are calculated as follows:

$$accuracy = \frac{\#\;of\;tokens\;predicted\;accurately}{\#\;of\;tokens}$$

\subsection{Evaluation}
The table \ref{tab:results1} presents the achieved results on both validation and test sets. For both punctuation and ezafe models, our proposed method achieved a higher F1 score regarding the related works in the fields. However, our proposed method performed second place for ZWNJ's. In the following, we will discuss each model separately.

\subsubsection{Punctuation}
The proposed method achieved an F1 score of 92.13\% in the test set. It is the first result in the punctuation restoration task using the Bijankhan dataset. However, we have found the work of \cite{Hosseini} with an F1 score of 69.00\% on a different dataset.
\subsubsection{Ezafe}
The proposed approach for ezafe recognition achieved the first F1 score of 98.50\% on the test set. Regarding obtained performance, the second best performing approach is \cite{ezafe1} that used XLMRoBERT and obtained an F1 score of  98.10\% with 0.40\% margin within our proposed model. In conclusion, the proposed model for ezafe recognitions is the best performer model.
\subsubsection{ZWNJ}
The work of \cite{zwnj11} used BERT-CRF with some additions to the training datasets. It boosted their performance from an F1 score of 96.67\% to 98.14\%. Nevertheless, in normal conditions, our proposed method performed 0.23\% higher. However, in the end, our model performed 2.24\% less than the highest approach in the task. However, they created a BIO structure while training their model but we used only a binary classification scheme and achieved F1 score of 96.90\%.

\begin{table}[]
    \centering
    \caption{Results}
    \begin{tabular}{c|c|c|c|c}
        \hline
        Model & Accuracy & Recall  & Precision & F1 \\
        \hline
        Validation\\
        \hline
        Punctuation & 98.59 & 91.10& 93.28& 92.09\\
        Ezafe & 98.89 &98.55 & 98.30& 98.42\\
        ZWNJ & 99.29& 98.86& 99.29& 96.92\\
        \hline
        Test\\
        \hline
        Punctuation & 98.60  & 90.13& 94.39& \textbf{92.13}\\
        Ezafe &98.94 &98.60 & 98.40& \textbf{98.50}\\
        ZWNJ &99.29 &96.71 & 97.09& 96.90\\
        \hline
    \end{tabular}
    \label{tab:results1}
\end{table}

\section{Conclusion}
In this paper, we presented ViraPart framework for automatic text refinements for various NLP tasks in the Persian language such as part-of-speech tagging, named entity recognition, ASR, grapheme to phoneme, spell checker, text summarization, and machine translation, etc. The ViraPart consists of 3 models, namely punctuation restoration, ZWNJ recognition, and ezafe recognition. We fine-tuned ParsBERT as a sequence labeling model for tasks and achieved promising results. We obtained averaged F1 macro score of 92.13\%, 98.50\%, and 96.90\% for punctuation restoration, ezafe recognition, and ZWNJ recognition, respectively. The presented results show the effectiveness of the proposed framework on text refinement.
\section*{Acknowledgment}
The authors would like to give big thanks to Part AI Research Center (the biggest AI company in Iran) for supporting and funding this research. 

\ifCLASSOPTIONcaptionsoff
  \newpage
\fi





\end{document}